\documentclass[runningheads]{llncs}
\usepackage{graphicx}

\usepackage{times}
\usepackage{soul}
\usepackage{url}
\usepackage[colorlinks=true,allcolors=blue]{hyperref}

\usepackage[utf8]{inputenc}
\usepackage[small]{caption}
\usepackage{subcaption}
\usepackage{graphicx}
\usepackage{amsmath}
\usepackage{amssymb}
\usepackage{booktabs}
\usepackage{algorithm2e}
\usepackage{xspace}
\usepackage[disable]{todonotes}
\usepackage{listings}
\usepackage[inline]{enumitem}
\newcommand{\dnot}{\ensuremath{\raise.17ex\hbox{\ensuremath{\scriptstyle\mathtt{\sim}}}}\xspace}                   
\newcommand{\pr}{\Pi}                
\newcommand{\gpr}{\Pi_G}
\newcommand{\cA}{\mathcal{A}}
\newcommand{\jss}{MPF-JSS\xspace}

\begin{document}
\title{Solving a Multi-resource Partial-ordering Flexible Variant of the Job-shop Scheduling Problem with Hybrid ASP}
\titlerunning{Solving MPF-JSS with Hybrid ASP}
%
\author{Giulia Francescutto\inst{1}\orcidID{0000-0002-4925-3346} \and
Konstantin Schekotihin\inst{2}\orcidID{0000-0002-0286-0958} \and
Mohammed M. S. El-Kholany\inst{2}\orcidID{0000-0002-1088-2081}}
\authorrunning{G. Francescutto et al.}
%
\institute{Infineon Technologies Austria AG, Siemensstrasse 2, 9500 Villach, Austria\\\email{giulia.francescutto@infineon.com} \and
University of Klagenfurt, Universitaetsstrasse 65-67, Klagenfurt, Austria
\email{\{konstantin.schekotihin,mohammed.el-kholany\}@aau.at}}
\maketitle              
\begin{abstract}
Many complex activities of production cycles, such as quality control or fault analysis, require highly experienced specialists to perform various operations on (semi)finished products using different tools. In practical scenarios, the selection of a next operation is complicated, since each expert has only a local view on the total set of operations to be performed. As a result,  decisions made by the specialists are suboptimal and might cause significant costs. 
In this paper, we consider a \emph{Multi-resource Partial-ordering Flexible Job-shop Scheduling} (\jss) problem where partially-ordered sequences of operations must be scheduled on multiple required resources, such as tools and specialists. The resources are flexible and can perform one or more operations depending on their properties.
The problem is modeled using Answer Set Programming (ASP) in which the time assignments are efficiently done using Difference Logic. Moreover, we suggest two multi-shot solving strategies aiming at the identification of the time bounds allowing for a solution of the schedule optimization problem. Experiments conducted on a set of instances extracted from a medium-sized semiconductor fault analysis lab indicate that our approach can find
schedules for 87 out of 91 considered real-world instances.

\end{abstract}

\section{Introduction}
Digitalization of manufacturing brings many advantages to the modern industry. 
Nevertheless, the work of highly experienced specialists cannot be substituted by machines in many fields like quality control, fault analysis, or research and development. 
In such scenarios, the experts use their knowledge of the application domain and apply sophisticated tools to perform a variety of operations on queued jobs.
In absence of any automated support, the specialists select jobs and perform operations that appear to be best according to their awareness of the situation.
That is, they are making locally best decisions according to numerous heuristics such as deadlines of the jobs, availability of tools and colleagues experienced in specific operations, or given preferences of jobs.
However, missed deadlines of jobs, as well as the idle time of machines and experts can be very costly.

In manufacturing settings, the reduction of the operational costs is often achieved by applying automated schedulers. \emph{Job-Shop Scheduling} (JSS) \cite{johnson1954optimal} is one of the most well-known problems in which, given a set of machines and a set of jobs represented as a sequence of operations, the goal is to assign operations to machines such that:
\begin{enumerate*}[label=\emph{(\roman*)}]
	\item each operation can be processed by one machine at a time;
	\item all jobs are processed independently of other jobs; and
	\item the execution of operations cannot be interrupted (no preemption).
\end{enumerate*}
Practical scheduling applications resulted in various extensions of JSS, such as \emph{flexible} JSS \cite{DBLP:journals/computing/BruckerS90} in which an operation can be performed by various resources, e.g., machines or engineers; \emph{multi-resource} JSS with \emph{non-linear routing} \cite{DBLP:journals/eor/Dauzere-PeresRL98}, where an operation may need multiple resources for its execution, e.g., an engineer and a machine, and may have different preceding and succeeding operations.  
Another well-known representation is the Resource Constrained Project Scheduling Problem (RCPSP) \cite{DBLP:journals/eor/HartmannB10}, where each operation might consume some amount of available resources. Similarly, as for JSS, there are various RCPSP extensions, such as a \emph{multi-skill} variant \cite{DBLP:journals/4or/Bellenguez-Morineau08}, where human resources might require skills to perform operations, or \emph{multi-mode} operations \cite{sprecher1997exact}, which can be performed in different ways, e.g., using different tools and procedures.

In this paper, we consider a Multi-resource Partial-ordering Flexible JSS (\jss) problem, which can informally be described as follows: given a set of jobs, represented as \emph{partially-ordered} sets of operations, and \emph{two sets of resources} that can \emph{perform multiple operations}, i.e., tools and engineers trained to operate them, find a schedule of operations for both machines and engineers that is optimal wrt.\ predefined criteria such as tardiness. The latter is defined as either $0$ if a job is done according to the computed schedule, or the difference between the completion time of this job and its deadline.
The partial order of operations indicates that the sequence of some operations of a job is not important. For instance, various inspection operations using different tools are non-invasive and can be done in an arbitrary order. Selection of a specific order can, however, improve the schedule since the availability of resources, like tools or engineers, might be limited. 

To solve the problem, we propose an encoding using Answer Set Programming (ASP) \cite{DBLP:books/sp/Lifschitz19} with Difference Logic \cite{DBLP:journals/tplp/JanhunenKOSWS17}. The introduction of difference constraints allows one to express timing requirements compactly and thus to avoid grounding issues that might occur if the number of possible time points required to find a schedule is too large. Nevertheless, as our evaluation shows, conventional reasoning and optimization methods of ASP solvers cannot find solutions to real-world instances in a predefined time. Therefore, we suggest two search strategies based on multi-shot solving techniques \cite{DBLP:journals/tplp/GebserKKS19}, allowing for the identification of tighter upper time bounds on the schedule. The evaluation of our approach was conducted on instances extracted from the historical data representing ten operational days of an Infineon Fault Analysis lab.
Each complete instance representing a whole day was then split into smaller instances enabling a detailed assessment of the solving performance. The results show that the basic ASP encoding was unable to solve any of the complete instances, while the suggested multi-shot approaches could find optimal schedules for eight or nine days, respectively. In total, these approaches solved 87 out of 91 instances considered in our full evaluation.


\section{Preliminaries}


\paragraph{Answer Set Programming.}\label{subsec:asp}
A normal ASP program $\Pi$ is a finite set of rules of the form%
\begin{equation}
	h \gets b_1, \ldots, b_m, \dnot b_{m+1}, \ldots, \dnot b_n
	\label{eq:rule}
\end{equation}
where $h$ and $b_1, \ldots b_n$, for $n \geq 0$, are atoms and $\dnot$ is \emph{negation as
failure}.
An \emph{atom} is an expression of the form $p(t_1, \ldots, t_l)$, where $p$ is a predicate symbol and $t_1, \ldots, t_l$ are \emph{terms}. 
Each term is either a \emph{variable} or a \emph{constant}.
A \emph{literal} $l$ is either an atom (positive) or its negation (negative).
Given a rule $r$ of the form \eqref{eq:rule}, the set $H(r) = \{h \}$ is the \emph{head} and the set $B(r) = B^+(r) \cup B^-(r) = \{b_1, \ldots, b_m \} \cup \{ \dnot b_{m+1}, \ldots, \dnot b_n \}$ 
is the \emph{body} of $r$, where $B^+(r)$ and $B^-(r)$ contain the
positive and negative body literals, respectively. 
A rule $r$ is a \emph{fact} if $B(r) = \emptyset$ and a \emph{constraint} if $H(r) = \emptyset$.
In addition, we denote the complement of a literal $l$ by $\overline{l}$ and by $\overline{L} = \{ \overline{l} \mid l \in L\}$ the complement for a set of literals $L$.
An atom, a literal, or a rule is \emph{ground} if no variables appear in it. 
A ground program $\gpr$ can be obtained from $\pr$ by substituting the variables in each rule $r \in \pr$ with constants appearing in $\pr$.

The semantics of an ASP program $\pr$ is given for its ground instantiation $\gpr$.
Let $\cA$ be the set of all ground literals occurring in $\gpr$. An
\emph{interpretation} is a set $I\subseteq \cA \cup \overline{\cA}$ of
literals that is \emph{consistent}, i.e., $I \cap \overline{I}=\emptyset$;
each literal $l \in I$ is true, each literal $l \in
\overline{I}$ is false, and any other literal is undefined.
An interpretation $I$ is \emph{total}, if $\cA\subseteq I\cup \overline{I}$.
An interpretation $I$ \emph{satisfies} a rule $r \in \gpr$, if $H(r)
\cap I \neq \emptyset$ whenever $B(r) \subseteq I$. A \emph{model} of $\gpr$ is
a total interpretation $I$ satisfying each $r\in \gpr$;
moreover, $I$ is \emph{stable} (an \emph{answer set}), if the set of atoms in
$I$ is $\subseteq$-minimal among all models of the reduct 
$\{H(r) \gets B^+(r) \mid r \in \gpr, B^-(r) \cap \overline{I} =
\emptyset\}$ \cite{DBLP:conf/iclp/GelfondL88}. 
Any answer set of $\gpr$ is also an answer set of $\pr$.

\paragraph{Multi-shot Solving.}\label{subsec:multi-shot}
ASP is a paradigm that introduces a flexible reasoning process that is suitable for controlled solving of continuously changing logic programs, i.e., multi-shot solving \cite{DBLP:journals/tplp/GebserKKS19}.
To accomplish this, \emph{clingo} enhances the ASP declarative language \cite{DBLP:journals/tplp/CalimeriFGIKKLM20} with control capacities. The former is achieved with the introduction of a new \lstinline{#program} directive in the ASP program that allows to structure it into subprograms, making the solving process completely modular. The latter is provided by an imperative programming interface that allows a continuous assembly of the program and gives control over the grounding and solving functions.
Each \lstinline{#program} subprogram has a name and an optional list of parameters. It gathers all the rules up to the next \lstinline{#program} directive. Subprogram \emph{base} is a dedicated subprogram where all the rules not preceded by any \lstinline{#program} directive are collected. 
\lstinline{#external} directives are used within subprograms to set external atoms to some truth value via the \emph{clingo} API.

\paragraph{ASP modulo Difference Logic.} 
\emph{clingo}[DL] extends the input language of \emph{clingo} by theory atoms representing difference constraints \cite{DBLP:conf/iclp/GebserKKOSW16,gebser_potassco_2019,DBLP:journals/tplp/JanhunenKOSWS17}. Difference constraints are represented by specific constraint atoms of the form 
$\text{\lstinline{&diff}}\{x-y\} \leq k$
where $x$ and $y$ are ASP terms, which are internally interpreted as integer variables' names, and $k$ is an integer. \emph{clingo}[DL] therefore provides the following extension of the normal rule~\eqref{eq:rule}:
\begin{align*}
    \text{\lstinline{&diff}}\{x-y\} \leq k \gets b_1, \ldots, b_m, \dnot b_{m+1}, \ldots, \dnot b_n.
\end{align*}
Such rules express that, whenever the body holds, the linear inequality represented by the head has to be satisfied as well.

\section{Problem Formalization}\label{sec:problemdef}

In this paper, we consider a novel variant of the JSS problem, which occurs in scenarios when   
multiple resources have to be combined in order to process incoming jobs. In particular, \jss extends the standard problem in three ways: 
\begin{enumerate*}[label=\emph{(\roman*)}]
	\item \emph{Multi-resource} -- there is more than one resource type needed to execute an operation;
	\item \emph{Partially-ordered} -- some operations of a job can be executed in an arbitrary order; and
	\item \emph{Flexible} -- an operation can be executed by various resources.
\end{enumerate*}

\subsection{\jss Definition}
\todo{Processing time for an operation is equal regardless of which machine it is using. We need to shift this to the demand in the future.}
Let $O=\{(o_1,p_1),\dots,(o_m,p_m)\}$ be a set of \emph{operations}, where $o_i$ denotes the operation identifier and $p_i \in \mathbb{N}$ its duration, and $C$ be a set of available \emph{classes of resources}, which represent groups of equivalent instances of a resource. 
Then, $R=\{(i,c,O_r) \mid i\in \mathbb{N}, c\in C, O_r \subseteq O\}$ is a set of available \emph{resources}, where each resource $r=(i,c,O_r)$ is a triple defining an \emph{instance} $i$ of the resource, its class, and a set of operations $O_r$ it can execute. 
In addition, the set $D=\{(o, C_d) \mid o \in O, C_d \subseteq C\}$ provides \emph{demands} of operations in $O$ 
for instances of resource classes $C_d$.
Finally, a set of \emph{jobs} is defined as $J=\{(O_j, P_j, d) \mid O_j \subseteq O, P_j \subseteq O_j\times O_j, d \in \mathbb{N}\}$, where $O_j \subseteq O$ is a set of operations that must be executed for the job, $P_j$ defines their (partial) order, and $d$ indicates the deadline.

Given an \jss instance $(J,D,R)$, a \emph{schedule} $S$ is a set of assignments $\{(R_s,o_j,t) \mid R_s \subseteq R, o_j \in O, t \in \mathbb{N}\}$. Each triple $(R_s,o_j,t)$ indicates that an operation $o_j$ of a given job and a required set of resources $R_s$ is assigned to a time point $t$. In addition, the following constraints must hold:
\begin{itemize}
	\item the set $R_s$ must comprise all resources demanded by an operation $o_j$;
	\item the schedule is non-preemptive, i.e., operations cannot be interrupted once started;
	\item any two operations of a job cannot be executed simultaneously; 
	\item each resource instance is assigned to only one operation at a time; and
	\item operations of a job $j$ must be scheduled wrt.\ to the given partial order, i.e., for any pair $(o_i, o_k) \in P_j$ the corresponding schedule assignments $(R_i,o_i,t_i)$ and $(R_k,o_k,t_k)$ must satisfy the inequality $t_{i} \le t_{k}$.
\end{itemize}
A schedule is \emph{optimal} if it has the minimal \emph{total tardiness} $T = \sum_{j \in J}\max(0,C_j-d_j)$, where $C_j$ and $d_j$ denote the completion time and the deadline of a job $j$, respectively.

\paragraph{Example.} 
Let us exemplify the \jss problem definition on a small instance. 
Suppose we have five operations $(o_1,1), \dots, (o_5,1)$ and two classes of resources -- a worker and a machine -- denoted by $w$ and $m$, respectively. 
The set of resources is defined as
\begin{align*}
	R=&\{(1,w,\{o_1,o_2\}),(2,w,\{o_4,o_5\}), (3,w,\{o_2,o_3,o_4\})\} \cup \\ 
	  &\{(1,m,\{o_3\}), (2,m,\{o_4\}), (3,m,\{o_4\}), (4,m,\{o_5\})\}
\end{align*}
For instance, in this set $(1,w,\{o_1,o_2\})$ indicates that the first worker is trained to execute operations $o_1$ and $o_2$, and $(1,m,\{o_3\})$ denotes that the first machine can be used to process $o_3$.
Moreover, the definition of the operation demand $D$ states that the operations $o_1$ and $o_2$ are processed only by workers, whereas the operations $o_3,o_4$ and $o_5$ require both a worker and a machine:
$$
	D=\{(o_1,\{w\}),(o_2,\{w\}), (o_3,\{w,m\}), (o_4,\{w,m\}), (o_5,\{w,m\})\}
$$

Assume that the shop got three new jobs with a deadline $3$ each. The first job has to undergo three operations, where $o_1$ must be completed before both $o_2$ and $o_4$. Operations of the second job can be done in an arbitrary order. Finally, the third job comprises four operations such that $o_3$ must be done before $o_2$ and $o_1$ before $o_2$ and $o_5$.
\begin{align*}
	J=\{&(\{o_1,o_2,o_4\},\{(o_1,o_2), (o_1,o_4)\}, 3),\\
	    &(\{o_3,o_4\},\emptyset, 3),\\ 
	    &(\{o_1,o_2,o_3,o_5\},\{(o_3,o_2), (o_1,o_2), (o_1,o_5)\}, 3)\}
\end{align*}
 
\begin{figure}[t]
	\centering
	\begin{minipage}{0.47\textwidth}
		\centering
		\includegraphics[width=\linewidth]{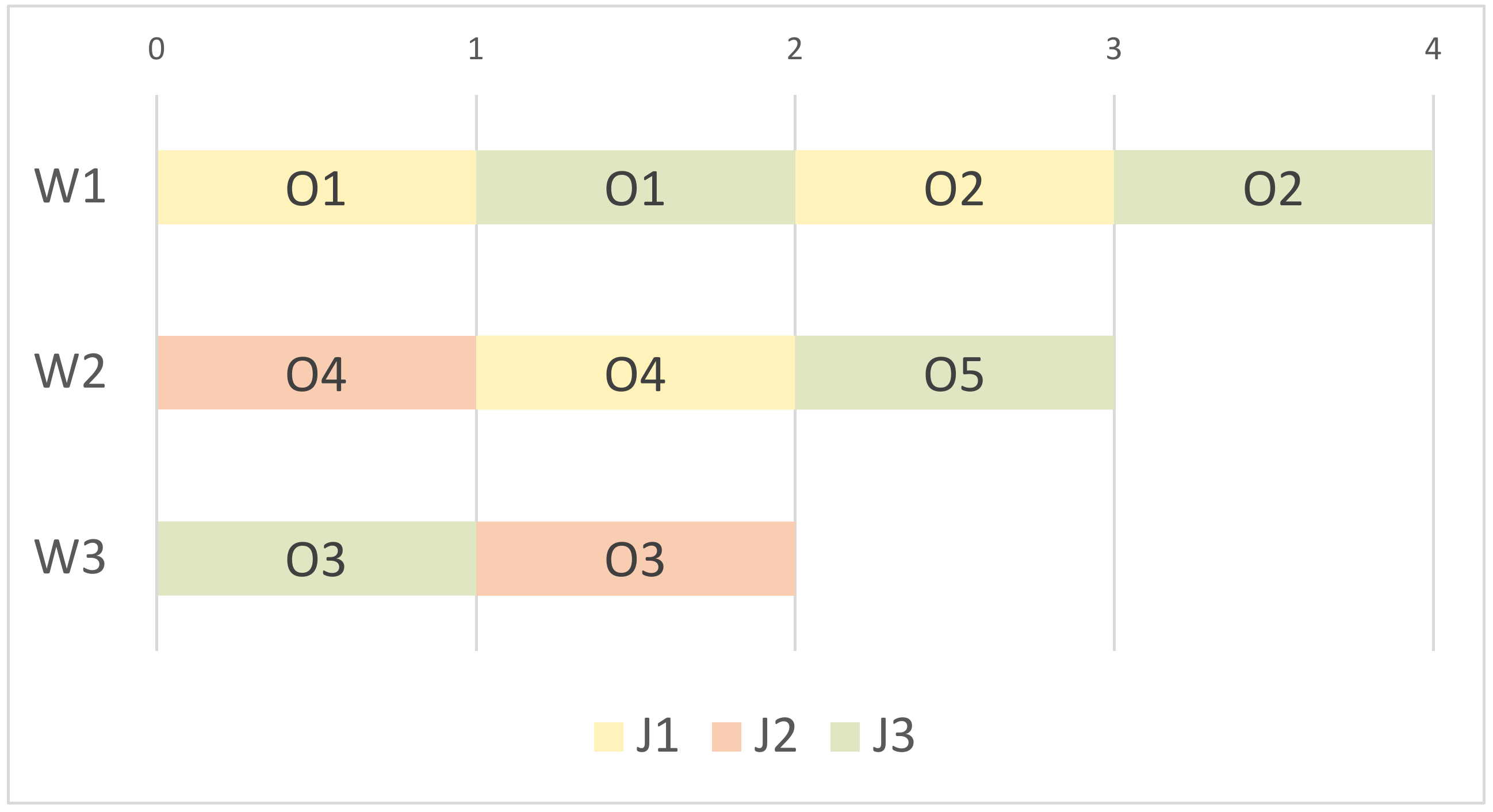}
		\caption{Workers allocations}
		\label{fig:w:alloc}
	\end{minipage}\hfill
	\hspace{.03\linewidth}
	\begin{minipage}{0.47\textwidth}
		\centering
		\includegraphics[width=\linewidth]{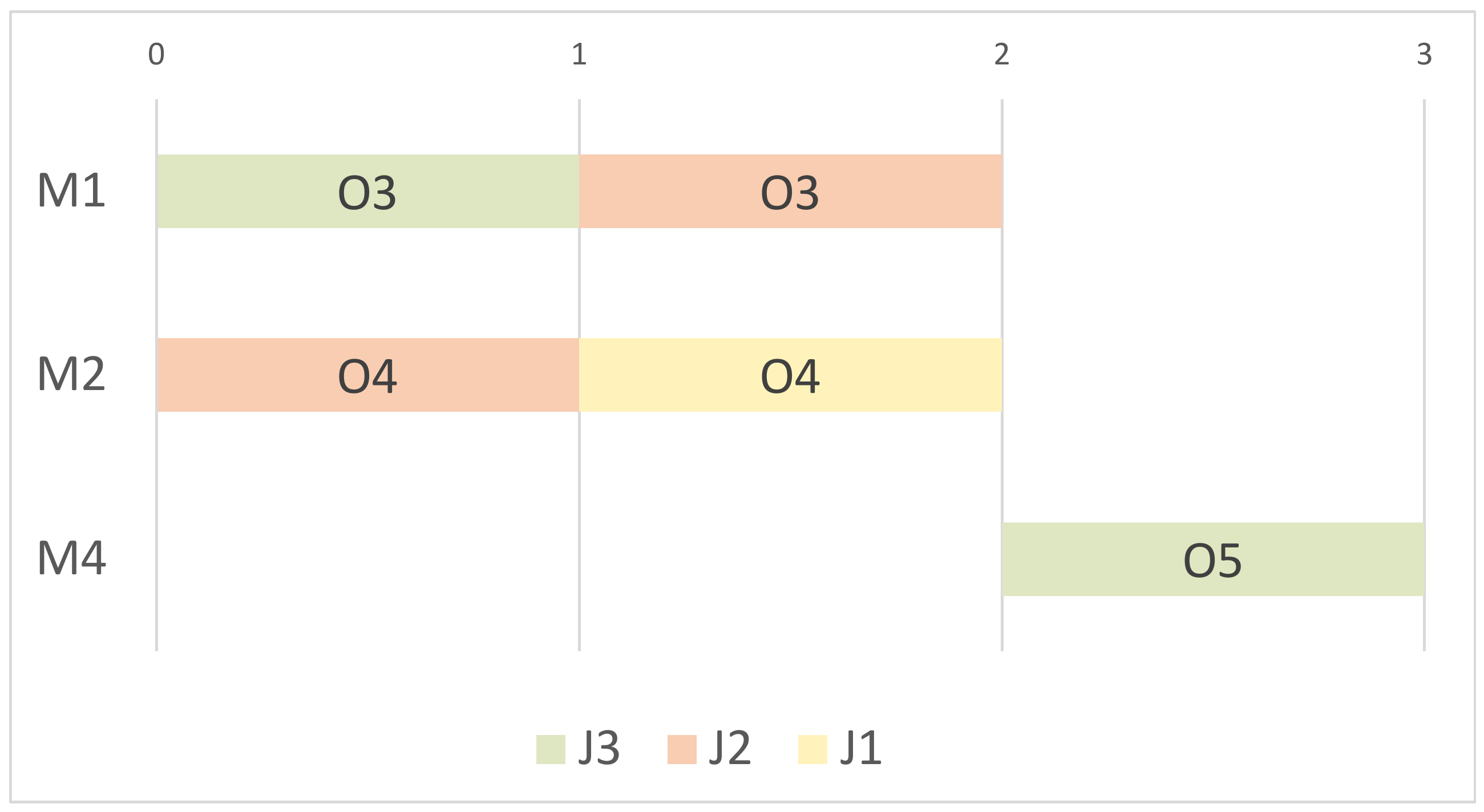} 
		\caption{Machines allocations}
		\label{fig:m:alloc}
	\end{minipage}
\end{figure}

One of the possible solutions of the given instance is shown in Figures \ref{fig:w:alloc} and \ref{fig:m:alloc}. The found schedule assigns operations to the provided resources and time points in a way that minimizes the tardiness optimization criterion. As a result, two jobs are finished in time, i.e., completion times of the first and second jobs are $C_1=3$ and $C_2=2$. The third job has tardiness $C_3-d_3=1$ since it is impossible to complete the four operations with a duration of $1$ each in time without executing operations of a job in parallel.

\subsection{Modeling \jss with Hybrid ASP}\label{sec:aspmodeling}
Answer Set Programming (ASP) has been widely used in the literature to solve scheduling problems. For instance, in \cite{DBLP:journals/corr/abs-1101-4554} ASP is used to develop a system for computing suitable allocations of personnel on the international seaport of Gioia Tauro. A similar problem of workforce scheduling is also addressed with ASP \cite{DBLP:conf/lpnmr/AbseherGMSW15}. In addition, ASP was applied to solve the course timetabling problem in \cite{DBLP:journals/tplp/BanbaraSTIS13} and \cite{DBLP:conf/padl/KahramanE19}. 
These approaches, however, indicated one of the major problems of ASP in scheduling applications -- grounding issues occurring while dealing with a large number of possible time points. 
Therefore, in \cite{DBLP:conf/lpnmr/Balduccini11}, ASP was integrated with Constraint Programming (CP) techniques to solve the problem of allocating jobs to devices in the context of industrial printing. A different integration of ASP with CP techniques was also suggested in \cite{10.1007/978-3-319-28697-6_23}, where two Constraint ASP approaches are proposed to solve a production scheduling problem. In this paper, we use ASP with Difference Logic to model the \jss problem, which was also applied in \cite{DBLP:conf/lpnmr/AbelsJOSTW19} to schedule railroad traffic.

\paragraph{Problem Instances.} 

In order to encode the \jss problem in ASP, we first define a number of predicates representing the input instances. 
The set of operations is encoded using the predicate \lstinline{op/2} where the first term is indicating the operation identifier, and the second the expected processing times. The demands of operations for resources are described with the \lstinline{needs/2} predicate, see lines \ref{prg:facts:ops:begin}-\ref{prg:facts:ops:end} in Listing \ref{prg:facts} encoding the example presented in the previous section.

The set of resources is represented with atoms over the \lstinline{res/3} predicate. An atom \lstinline{res(c,r,o)} provides a class \lstinline{c} of the required resource, an identifier \lstinline{r} of a resource instance, and an operation \lstinline{o} it can execute. Thus, the set of required resources can be encoded as shown in lines \ref{prg:facts:workers:begin}-\ref{prg:facts:machines:end}. 


Finally, the jobs are encoded using three predicates \lstinline{job/2}, \lstinline{recipe/2}, and \lstinline{prec/3}. Atoms over the first predicate provide identifiers of the jobs and their deadlines. Recipes are used to define the set of operations that must be executed for a job, and the partial order of the operations is specified by the atoms over the \lstinline{prec/3} predicate. Respective facts encoding the jobs of our example are given in lines \ref{prg:facts:jobs:begin}-\ref{prg:facts:jobs:end}.

\begin{lstlisting}[float=bt,mathescape=true,escapeinside={\#(}{\#)},basicstyle={\ttfamily\small},label=prg:facts,caption={Problem instance}, linerange={1-14} ]
op(o1,1). op(o2,1). op(o3,1). op(o4,1). op(o5,1).#(\label{prg:facts:ops:begin}#)
needs(o1,w). needs(o2,w). needs(o3,m). needs(o3,w).
needs(o4,m). needs(o4,w). needs(o5,m). needs(o5,w).  #(\label{prg:facts:ops:end}#) 

res(w,w1,o1). res(w,w1,o2). res(w,w2,o4). res(w,w2,o5).#(\label{prg:facts:workers:begin}#)
res(w,w3,o2). res(w,w3,o3). res(w,w3,o4).#(\label{prg:facts:workers:end}#) 
res(m,m1,o3). res(m,m2,o4). res(m,m3,o4). res(m,m4,o5).#(\label{prg:facts:machines:begin}#) #(\label{prg:facts:machines:end}#)  

job(j1,3). job(j2,3). job(j3,3). #(\label{prg:facts:jobs:begin}#)  
recipe(j1,o1). recipe(j1,o2). recipe(j1,o4).
recipe(j2,o3). recipe(j2,o4). 
recipe(j3,o1). recipe(j3,o2). recipe(j3,o3). recipe(j3,o5).         
prec(j1,o1,o2). prec(j1,o1,o4).
prec(j3,o3,o2). prec(j3,o1,o2). prec(j3,o1,o5).#(\label{prg:facts:jobs:end}#)
\end{lstlisting}

\paragraph{\jss Encoding.} 

The problem encoding is split into three parts: 
\begin{enumerate}
	\item \emph{base}: encodes all the definitions and constraints for the scheduling requirements (Listings \ref{prg:base:alloc}, \ref{prg:base:prec}, and \ref{prg:base:diffconst});
	\item \emph{incremental}: implements an incremental search strategy as well as weak constraints for the tardiness optimization (Listing \ref{prg:step}); and
	\item \emph{exponential}: uses multi-shot solving to find the upper bound on the tardiness for a given instance using exponential search (Listing \ref{prg:exp}).
\end{enumerate} 

The first section of the \emph{base} subprogram, presented in Listing \ref{prg:base:alloc}, addresses the allocation of resources, expressed by atoms \lstinline{alloc(R,J,O,M)}, required to execute an operation of a job. Each operation \lstinline{O} of a job \lstinline{J} requiring a resource of type \lstinline{R} should be executed by exactly one instance \lstinline{M} of this resource. Since the instances of a resource are equivalent, we introduce a symmetry-breaking constraint in lines \ref{prg:base:symm:begin}-\ref{prg:base:symm:end}. This constraint avoids unnecessary allocation variants by requiring the solver to select resources starting from the ones with the lexicographically smallest identifier.

\begin{lstlisting}[float=t,mathescape=true,escapeinside={\#(}{\#)},basicstyle={\ttfamily\small},label=prg:base:alloc,caption={Encoding of the resource allocation}, linerange={7-10} ]
{alloc(R,J,O,M) : res(R,M,O)}=1 :- recipe(J,O), needs(O,R).

allocated(R,M) :- alloc(R,J,O,M).  #(\label{prg:base:symm:begin}#)
:- alloc(R,J,O,N), res(R,M,O), M < N, not allocated(R,M). #(\label{prg:base:symm:end}#)
\end{lstlisting}

Listing \ref{prg:base:prec} shows the second part of the \emph{base} subprogram. The rule in  lines \ref{prg:base:order:begin}-\ref{prg:base:order:end} specifies that the order in which two operations of a job are executed can be arbitrary when these operations are not subject to the job's precedence relation. Similarly, an arbitrary order is possible if two operations of different jobs require the same resource (lines \ref{prg:base:order:res:begin}-\ref{prg:base:order:res:end}).
For the operations that need an ordering, expressed by atoms over the \lstinline{ord/4} predicate, we generate an execution sequence -- denoted by the \lstinline{seq/4} predicate -- using the rules in lines \ref{prg:base:seq:begin} and \ref{prg:base:seq:end}. 
The sequence of operations whose precedence is given in the input instance is forced by the rule in line \ref{prg:base:seq:prec}.

\begin{lstlisting}[float=t,literate={\%\%}{}{0},mathescape=true,escapeinside={\#(}{\#)},basicstyle={\ttfamily\small},label=prg:base:prec,caption={Encoding of operation sequences},firstnumber=6, linerange={12-19}]
ord(J,O1,J,O2)    :- recipe(J,O1), recipe(J,O2), O1 < O2,#(\label{prg:base:order:begin}#)
                     not prec(J,O1,O2), not prec(J,O2,O1). #(\label{prg:base:order:end}#)
ord(J1,O1,J2,O2)  :- alloc(R,J1,O1,M), alloc(R,J2,O2,M),#(\label{prg:base:order:res:begin}#)
                     J1 < J2. #(\label{prg:base:order:res:end}#)
{seq(J1,O1,J2,O2)}:- ord(J1,O1,J2,O2).#(\label{prg:base:seq:begin}#)
seq(J2,O2,J1,O1)  :- ord(J1,O1,J2,O2), not seq(J1,O1,J2,O2). #(\label{prg:base:seq:end}#)
seq(J,O1,J,O2)    :- prec(J,O1,O2). #(\label{prg:base:seq:prec}#)
\end{lstlisting}
\begin{lstlisting}[float=t,mathescape=true,escapeinside={\#(}{\#)},basicstyle={\ttfamily\small},caption={Difference constraints on operations' starting times}, firstnumber = 13,label=prg:base:diffconst, linerange={20-21}]
&diff{0 - (J,O)}<=0 :- recipe(J,O). #(\label{prg:base:diff_const:begin}#)
&diff{(J1,O1) - (J2,O2)}<=-P1 :- seq(J1,O1,J2,O2), op(O1,P1). #(\label{prg:base:diff_const:end}#)
\end{lstlisting}

Finally, in Listing \ref{prg:base:diffconst} we introduce the difference constraints encoding the starting times of operations. We represent the starting time of an operation \lstinline{O} of job \lstinline{J} by an integer variable \lstinline{(J,O)}. The first constraint in line \ref{prg:base:diff_const:begin} requires the starting time of each operation to be greater or equal to $0$. The second constraint enforces the starting times to be compatible with the order provided by atoms over the \lstinline{seq/4} predicate. That is, an operation \lstinline{(J2,O2)} coming after \lstinline{(J1,O1)} must not start before \lstinline{(J1,O1)} is finished.

\paragraph{Multi-shot Solving.} 
Finding solutions of minimal tardiness for \jss instances can be hard since, without the knowledge of any reasonable bounds on the scheduling time interval, a solver may have to enumerate a large number of possible solutions.
Finding such bounds can be complicated and simple heuristics, like determining a maximal sum of operation durations for a particular resource, often provide very imprecise approximations.  
Therefore, we exploit the power of multi-shot solving to find an upper bound on the tardiness and thus provide a good starting point for the optimization methods of an ASP solver.
In the following, we present two approaches to incrementally search for feasible solutions to a given problem instance.

\begin{figure}[b]
	\centering
	\includegraphics[width=0.7\linewidth]{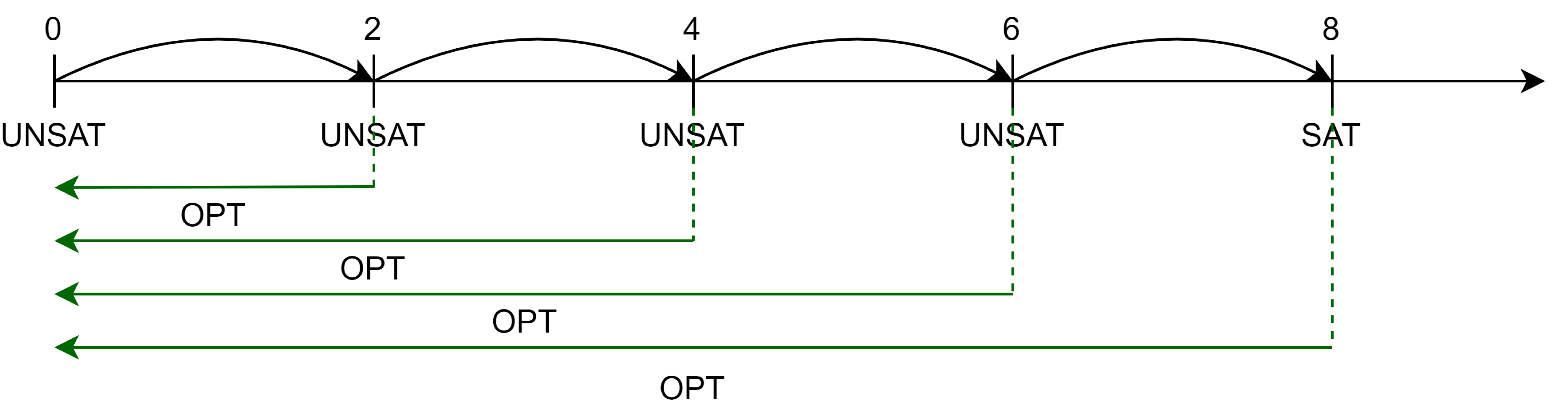}
	\caption{Incremental approach \label{fig:inc}}
	\vspace{5pt}
	\includegraphics[width=0.7\linewidth]{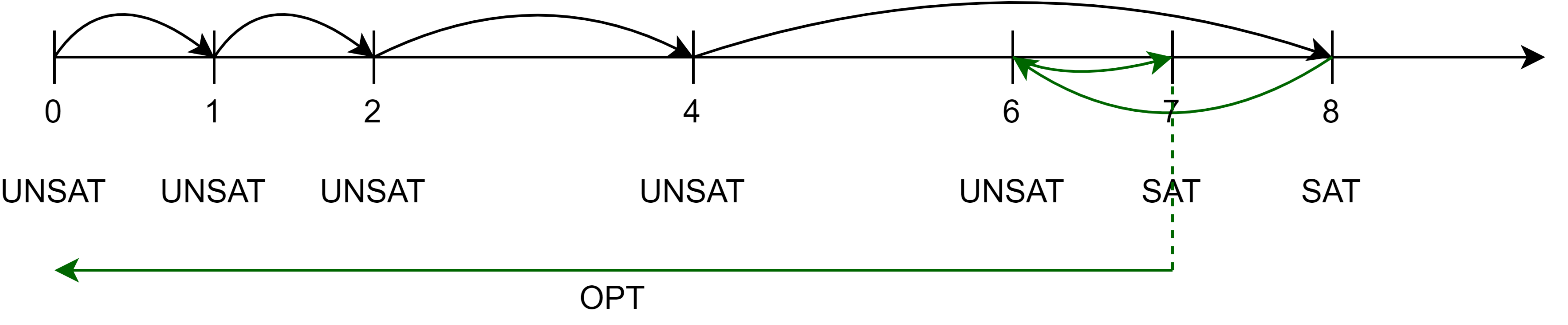}
	\caption{Exponential approach \label{fig:exp}}
\end{figure}

The idea of the first approach is to incrementally increase the upper bound on the tardiness of each job in order to identify an interval for which a schedule exists. The algorithm starts by considering the $0$ tardiness bound, and if this yields unsatisfiability (UNSAT), it starts to increment the tardiness bounds by a constant. As a result, the algorithm implements a tumbling window search strategy. The corresponding subprogram \lstinline{step(m,n)}, shown in Listing \ref{prg:step}, takes the parameters \lstinline{m} and \lstinline{n} to indicate the lower and upper bounds of the interval considered in the current iteration. The parameter values are set via a Python control script, which shifts them by the considered window size in each iteration. 
The control is implemented using the \lstinline{#external} directive, which provides a mechanism to activate or deactivate constraints by assigning a corresponding \lstinline{tardiness(n)} atom to \emph{true} or \emph{false}, respectively. 
Figure \ref{fig:inc} illustrates a sample execution of the incremental search algorithm, where a tumbling window of size $2$ is moved in each iteration until the target interval for which a schedule exists is found. 

Once the admissible upper bound \lstinline{n} (and a corresponding lower bound \lstinline{m}) is identified, ASP optimization methods search for an optimal solution within this interval,
where the truth of an \lstinline{end(J,N)} atom for \lstinline{N} in-between $1$ and \lstinline{n} expresses that the tardiness of job \lstinline{J} is less than \lstinline{N}.
Such atoms can be guessed to be \emph{true} via the choice rule in line \ref{prg:step:begin}. The constraint in line \ref{prg:step:tardiness:one} propagates smaller tardiness up to the upper bound~\lstinline{n}, and line \ref{prg:step:tardiness:two} forces the tardiness of each job to be less than \lstinline{n}. Finally, we minimize the number of pairs \lstinline{J,N} for which \lstinline{end(J,N)} is \emph{false}, i.e., the tardiness of job \lstinline{J} is at least~\lstinline{N}, by means of the weak constraint in line \ref{prg:step:end}. 
Such an optimization strategy is required since 
\emph{clingo}[DL] does not directly allow for minimizing a sum of integer variables occurring in its difference constraints. 
Therefore, in lines \ref{prg:step:constraint:begin}-\ref{prg:step:constraint:end}, we force each operation of a job to finish within the corresponding tardiness bound. 
In an obtained answer set, the \lstinline{end(J,N)} atom with the smallest value for \lstinline{N} signals the tardiness \lstinline{N-1} for job \lstinline{J}.

\begin{lstlisting}[float=t,mathescape=true,escapeinside={\#(}{\#)},basicstyle={\ttfamily\small},caption={Step subprogram for incremental approach}, firstnumber = 16,label=prg:step, linerange={33-44}]
#program step(m,n). 
#external tardiness(n).

current(m..n,n).

{end(J,N)} :- job(J,_), current(N,n).#(\label{prg:step:begin}#)
:- end(J,N-1), not end(J,N), current(N,n).#(\label{prg:step:tardiness:one}#)
:- job(J,_), not end(J,n), tardiness(n).#(\label{prg:step:tardiness:two}#)
:~ job(J,_), not end(J,N), current(N,n). [1,J,N]#(\label{prg:step:end}#)

&diff{(J,O) - 0}<=K :- job(J,D), end(J,N), current(N,n), #(\label{prg:step:constraint:begin}#)
                       recipe(J,O), op(O,P), K=D+N-P. #(\label{prg:step:constraint:end}#)
\end{lstlisting}

In the second approach, shown in Listing \ref{prg:exp}, the additional subprogram \lstinline{iterate(n)} is used first to find an upper tardiness bound \lstinline{n} for each job such that some schedule exists.
This is accomplished by a binary search that exponentially increments \lstinline{n} until the first schedule is found, and then converges to the smallest \lstinline{n} for which the scheduling problem is still satisfiable (SAT).
Figure \ref{fig:exp} illustrates the process converging to the upper bound~$7$,
relative to which the tardiness optimization is performed in the second step.
In fact, with the \lstinline{iteration(n)} atom from the \lstinline{#external} directive in line~\ref{prg:iterate:external} set to \emph{true},
the constraint in lines \ref{prg:iterate:begin}-\ref{prg:iterate:end} forces the
tardiness of each job to be less than~\lstinline{n},
and it remains to add the \lstinline{step(1,n-1)} subprogram as above
for optimization, yet letting the external atom \lstinline{tardiness(n-1)}
be \emph{false} to avoid unsatisfiability due to the constraint in line~\ref{prg:step:tardiness:two}.
%

\begin{lstlisting}[float=tb,mathescape=true,escapeinside={\#(}{\#)},basicstyle={\ttfamily\small},caption={Iterate subprogram for exponential approach}, firstnumber = 29,label=prg:exp, linerange={23-27}]
#program iterate(n).
#external iteration(n).#(\label{prg:iterate:external}#)

&diff{(J,O) - 0}<=K :- job(J,D), iteration(n),#(\label{prg:iterate:begin}#)
                       recipe(J,O), op(O,P), K=D+n-P.#(\label{prg:iterate:end}#)
\end{lstlisting}

\paragraph{Discussion.} The two presented approaches aim at finding tardiness bounds per job, which can then be used as a starting point for the tardiness optimization. 
However, the bounds found by both approaches might have a different impact on the total tardiness of the optimal solution. 
This difference is essential for the quality of the obtained solutions, and both bounds merely approximate the maximal tardiness of some jobs needed for the optimal schedule. Theoretically, there can be situations in which the minimal total tardiness can only be reached when one of the jobs has comparably high tardiness, whereas others can be completed with low or zero tardiness. The upper bounds identified by our search algorithms might result in suboptimal schedules in which jobs, nevertheless, do not have extreme differences in their tardiness. In particular, the exponential approach is geared to find solutions that avoid large tardiness differences between jobs. Such schedules might be advantageous in scenarios where customers can tolerate short waiting times without serious drawbacks. The incremental strategy may admit schedules with smaller total tardiness, given a sufficiently large window size to come to a greater bound than identified by the exponential search. However, the incremental approach can be computationally more expensive since it might result in a larger number of candidate solutions to be considered by the underlying ASP optimization method.


\section{Experimental Evaluation}

We conduct our experiments on a set of real-world instances of the \jss problem retrieved from the daily operations history of a semiconductor Fault Analysis (FA) Lab.  

\paragraph{Application Domain: FA Lab.}
In the context of semiconductor industries, the goal of the FA process is to identify the failure that results in an observed incorrect behavior of a semiconductor device \cite{committee2011microelectronics}. In order to determine the nature and the cause of the failure, a sequence of investigation activities must be performed. Thus, different FA methods are applied to correctly identify different aspects of the failure. Obtained results are then put together to infer the failure mechanism and to understand its causes.

Some FA techniques alter the device permanently, e.g., the chemical alteration of the surface, while others do not affect the device in any way. Consequently, some analyses need to be executed before any alteration of a sample semiconductor device, while others can only be executed after some specific alteration. For example, an initial external visual inspection can only be done on a non-altered device, whereas the internal inspection of the sample is possible only after its decapsulation. This results in some precedence requirements for the sequences of executed investigation methods.\footnote{\url{www.eesemi.com}}

In general, a sequence of FA techniques, required to identify the root cause of a fault, is unknown. Often the next method is selected upon results obtained during the previous steps of the FA process. However, in the specific context we consider in this paper, the FA process is executed to assess the quality of  produced devices that have been put under some stress test. Therefore, in this case the set of methods to be executed is known in advance.  

Each FA technique is executed by a trained employee possibly using a dedicated machine. For most of the operations, there is more than one employee trained to conduct them using one of the available machines. 
FA diagnostics of all incoming devices, called jobs, must be finished within a certain predefined deadline. The latter might not be met due to unavoidable reasons, like personnel shortages, unavailability of tools, or large numbers of incoming jobs. However, in practice, most of the missed deadlines are due to ad-hoc scheduling that leads to ``forgotten'' jobs, bottlenecks by rare and expensive equipment, or personnel allocation.
One of the possible solutions is, therefore, to represent the FA problem as \jss  and use modern ASP solvers to find the optimal allocation of jobs' analyses to machines and employees. 

\paragraph{Instance Generation.}
The experimental evaluation is performed on real-world instances provided by our partner Infineon Technologies Austria. 
The provided data comprises a list of jobs processed in a selected period of time including: 
\begin{enumerate*}[label=\emph{(\roman*)}]
	\item the list of operations/techniques executed on each job, 
	\item the processing time of the operations, 
	\item the job deadlines, 
	\item job identifiers, 
	\item the list of machines available, and 
	\item information about the employees who executed these operations. 
\end{enumerate*}

Out of this data, we extracted instances for ten random days representing a snapshot of the situation in the lab. However, the information regarding precedence requirements between failure analysis techniques was missing in the provided data. At the moment of data collection, it was impossible to communicate to the experts and to retrieve the missing dependencies between the techniques. Therefore, we decided to take the order of operations given in the data and convert it into an ordering relation, thus, providing a strict order in our test instances.\footnote{See the \href{https://git-ainf.aau.at/Giulia.Francescutto/papers/-/wikis/jelia20}{paper website} for the encodings and instances.}
The processing time of each operation was set to its average processing time measured in minutes during the selected period. Given the job deadlines in a date format, we computed the difference in minutes between the deadline date and the beginning of the day shift chosen for the computation. For simplicity, we assumed that each day has only one shift, which is eight hours long. 

The resulting instances have the following approximate number of fixed components defined by the properties of the studied lab:
\begin{enumerate*}[label=\emph{(\roman*)}]
	\item $50$ operations,
	\item $75$ machines, and 
	\item $45$ workers.
\end{enumerate*}
For each of the days chosen the number of open jobs ranges between $30$ and $50$. We split each day-instance into sub-instances, with the aim of having instances of increasing size in multiples of $5$, which resulted into a total of $91$ instances.

\paragraph{Evaluation Results.}
For each obtained instance, we ran the three ASP programs introduced in Section \ref{sec:aspmodeling}: 
\begin{enumerate*}[label=\emph{(\roman*)}]	
	\item \emph{single-shot} -- the base program with a tardiness bound precomputed using a heuristic,
	\item \emph{inc} -- the incremental variant, and 
	\item \emph{exp} -- the exponential search approach.
\end{enumerate*}
We compared the multi-shot approaches to a single-shot version where the bound is computed in advance and given as constant input to the program. In particular, we compute the sum of durations of all operations in an instance, which defines the tardiness bound large enough to allow for finding an optimal solution. 
Then, we introduce this bound as constant in the ASP single-shot program, and solve the optimization problem using the same idea as for the optimization steps of the multi-shot approaches.

The experiments were conducted on a workstation with Ubuntu $18.05$, Intel  $3930$K and $64$GB RAM. In our experiments, we use \emph{clingo}[DL] 1.1.0 and \emph{clingo} 5.4.0 with multi-shot solving controlled by the main routine in Python 3.8.5. For each instance, we let the solver run up to a timeout of $2$ hours. For the incremental approach, we chose to use a constant tumbling window of $20$ minutes.

\begin{figure}[b]
	\centering
	
	\includegraphics[width=0.75\linewidth]{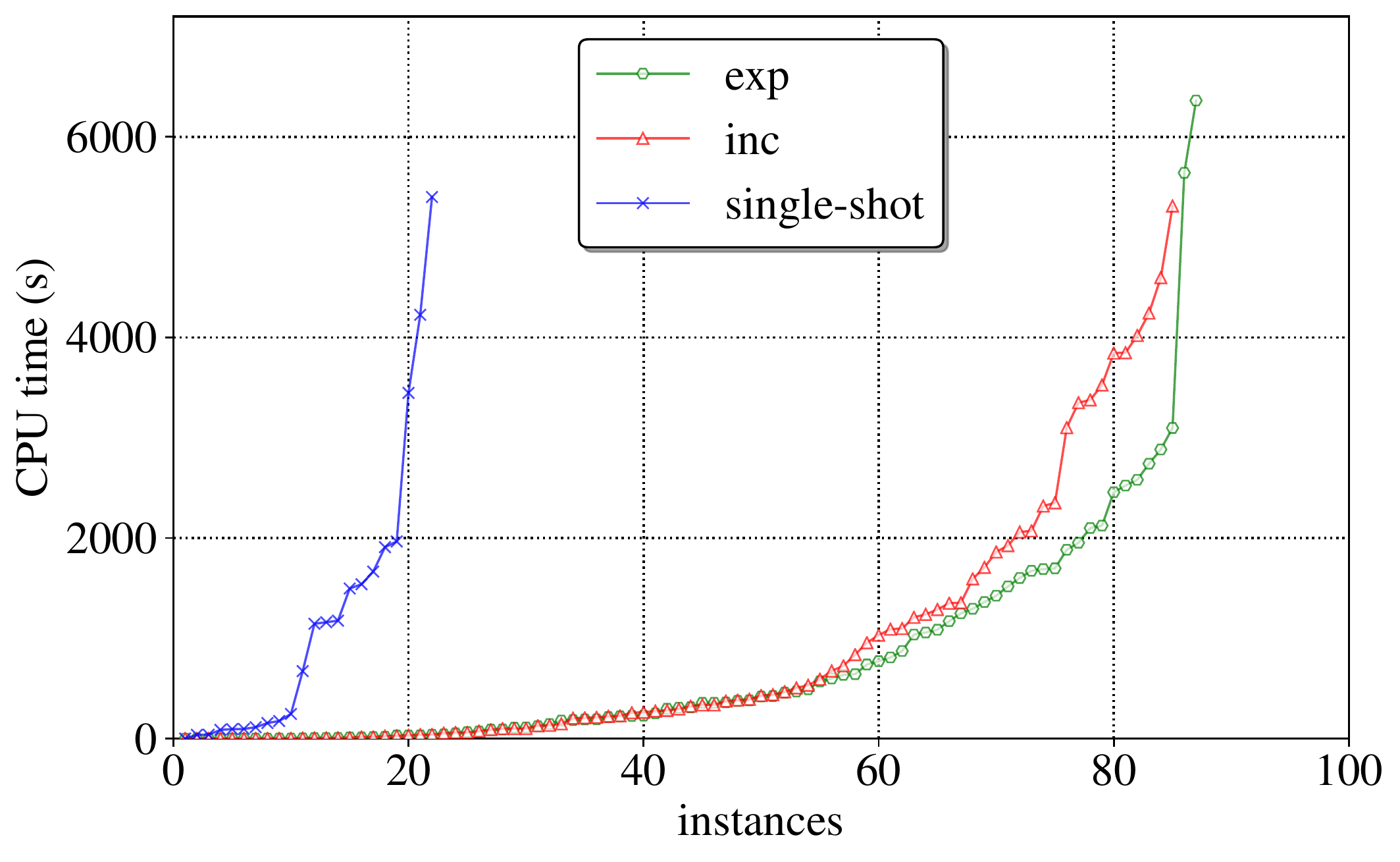}
	\caption{Cactus plot of solving times \label{fig:cactus:time}}
	  
	\vspace{5pt}
	
	\includegraphics[width=\linewidth]{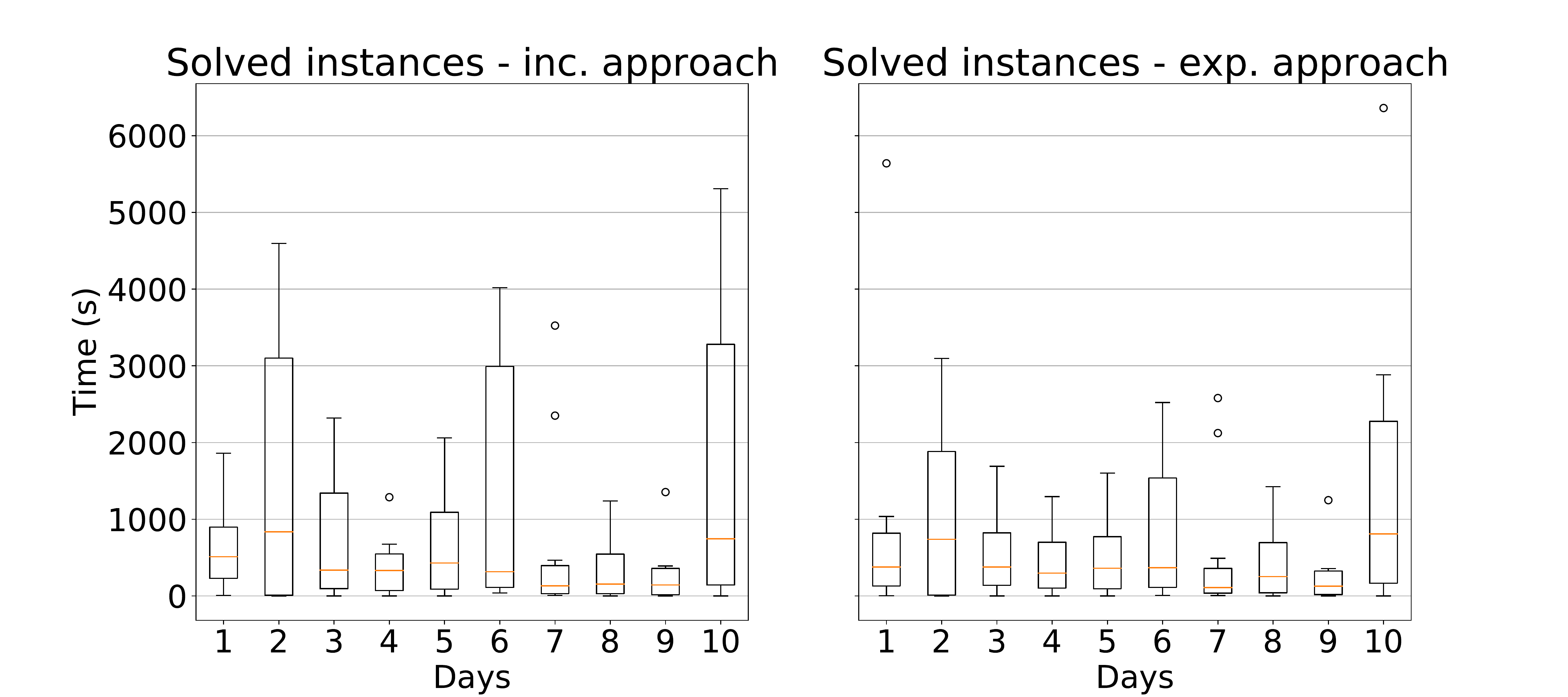}
	\caption{Box plot of solving times for the instances from ten days and the two multi-shot approaches. The incremental approach reached the timeout for five instances in Day 1 and one instance in Day 10, and the exponential approach reached the timeout for four instances in Day~1.\label{fig:boxplot}}
\end{figure}

Figure \ref{fig:cactus:time} shows a comparison of the solving performance of the two multi-shot approaches and the single-shot version. The single-shot program manages to solve only a small subset of the test instances. Thus, it always reached the timeout for instances with more than $20$ jobs and only managed to find a schedule for instances with $15$ jobs for two days -- Day 2 and Day 9.
Interestingly, the total tardiness of all schedules found by the single-shot and by multi-shot approaches was equal.
The two multi-shot approaches significantly outperformed the single-shot version. The exponential version solved $87$ instances and reached the timeout only for four instances, all of which belong to Day~$1$. The incremental version managed to solve $85$ instances and reached the timeout for six instances: five instances in Day 1, and the largest instance in Day 10 comprising all recorded jobs for this day. 
The exponential approach performs slightly better since it was always finding a tighter upper bound for tardiness, thus, leaving fewer choices for the optimization strategy of the ASP solver. Nevertheless, the differences between the approaches discussed in Section \ref{sec:aspmodeling} were also confirmed in the evaluation. That is, in our experiments, the incremental variant was able to obtain better solutions for three instances with an average total tardiness improvement of 80 minutes.

The box plot in Figure \ref{fig:boxplot} summarizes the solving times measured over instances from each day. We observe that the exponential approach generally needs less time. The outliers in the plot of the exponential search correspond to instances that the incremental version did not manage to solve within the timeout. Nevertheless, if we consider only the solved instances, then the solving times required by both multi-shot approaches are quite close with a slight advantage of the exponential search. The reason is that \emph{clingo}[DL] could rather quickly decide whether an instance is satisfiable for a given tardiness bound or not. For the largest instances, the exponential and incremental strategies required 75 and 211, respectively, seconds on average for finding a bound, while substantially more time was spent on the optimization performed wrt.\ this bound.

Table \ref{tab:times} shows the times needed by the two multi-shot approaches for their search and optimization steps on the largest instance per day. In general, the exponential approach is faster to find a tardiness bound than the incremental strategy.
In Day 1, both approaches reached the timeout of $7200$ seconds in the optimization step, and in Day 10 only the exponential approach managed to find an optimal solution within the timeout. 

\begin{table}[tb]
	\centering
\begin{tabular}{|l|r|r|r|r|r|r|r|r|r|r|}
	\hline
	{} &  Day1 &  Day2 &  Day3 &  Day4 &  Day5 &  Day6 &  Day7 &  Day8 &  Day9 & Day10 \\
	\hline
	inc\_search &   362 &   285 &   174 &   131 &   140 &   150 &   165 &   251 &   281 &   173 \\
	inc\_opt    &  TO &  4311 &  2145 &  1156 &  1920 &  3869 &  3360 &   987 &  1074 &  TO \\ \hline
	exp\_search &   151 &    77 &    49 &    39 &    75 &    47 &    52 &    41 &    15 &   208 \\
	exp\_opt    &  TO &  3020 &  1640 &  1256 &  1442 &  2476 &  2528 &  1384 &  1233 &  6154 \\
	\hline
	\end{tabular}
	\caption{Search and optimization times in seconds for multi-shot approaches on the largest instances}
	\label{tab:times}
\end{table}

\section{Conclusions}
In this paper, we introduce a Multi-resource Partial-ordering Flexible Job-Shop Scheduling (\jss) problem
and provide an encoding using hybrid ASP with Difference Logic. We present two multi-shot solving strategies to find reasonable approximations of tardiness bounds. These approaches were tested on a set of real-world instances provided by Infineon Technologies Austria, where they showed to enable the optimization of daily schedules for a Fault Analysis lab,
while single-shot solving could not accomplish the optimization within the same time limit.

In the future, we are going to extend the suggested approach in two ways. First, we intend to develop novel optimization techniques for Difference Logic allowing for the minimization of sums of integer variables, which must in the current approach be simulated by means of underlying ASP optimization methods.  
Second, we aim to devise multi-shot solving strategies that can take advantage of historical data, which is often available in industrial application scenarios. 
In particular, we are going to study combinations of ASP with (supervised) machine learning models trained to guide the search procedures of a solver by giving preference to operations to be scheduled in successive solving steps. 

%
%

\bibliographystyle{splncs04}
\bibliography{mpfjss}

\end{document}